\documentclass[letterpaper]{article} 
\usepackage{aaai2026}  
\usepackage{times}  
\usepackage{helvet}  
\usepackage{courier}  
\usepackage[hyphens]{url}  
\usepackage{graphicx} 
\usepackage{natbib}  
\usepackage{caption} 
\usepackage{algorithm}
\usepackage{algorithmic}

\usepackage{array} 
\usepackage{booktabs} 
\usepackage{amsmath} 
\usepackage{multirow} 
\usepackage{amssymb}
\usepackage{url}
\usepackage{newfloat}
\usepackage{listings}
\DeclareCaptionStyle{ruled}{labelfont=normalfont,labelsep=colon,strut=off} 

\floatstyle{ruled}
\newfloat{listing}{tb}{lst}{}
\floatname{listing}{Listing}
\title{AAAI Press Formatting Instructions \\for Authors Using \LaTeX{} --- A Guide}
\author{
    Written by AAAI Press Staff\textsuperscript{\rm 1}\thanks{With help from the AAAI Publications Committee.}\\
    AAAI Style Contributions by Pater Patel Schneider,
    Sunil Issar,\\
    J. Scott Penberthy,
    George Ferguson,
    Hans Guesgen,
    Francisco Cruz\equalcontrib,
    Marc Pujol-Gonzalez\equalcontrib
}
\affiliations{
    \textsuperscript{\rm 1}University of Southampton\\
    \textsuperscript{\rm 2}MogoAI

    fudongjie@mogoace.com, ts1v23@soton.ac.uk, p.fang@soton.ac.uk, x.cai@soton.ac.uk, H.Kim@soton.ac.uk

}

\title{MOGO: Residual Quantized Hierarchical Causal Transformer for Real-Time and Infinite-Length 3D Human Motion Generation}

\author{
  Dongjie Fu\textsuperscript{\rm 2, \equalcontrib},
  Tengjiao Sun\textsuperscript{\rm 1,2,\equalcontrib},
  Pengcheng Fang\textsuperscript{\rm 1,2, \equalcontrib}
  Xiaohao Cai\textsuperscript{\rm 1}
  Hansung Kim\textsuperscript{\rm 1, \thanks{Corresponding author.}}
}

\usepackage{bibentry}

\begin{document}

\maketitle
\begin{abstract}

Recent advances in transformer-based text-to-motion generation have significantly improved motion quality. However, achieving both real-time performance and long-horizon scalability remains an open challenge. In this paper, we present MOGO (Motion Generation with One-pass), a novel autoregressive framework for efficient and scalable 3D human motion generation. MOGO consists of two key components. First, we introduce MoSA-VQ, a motion scale-adaptive residual vector quantization module that hierarchically discretizes motion sequences through learnable scaling parameters, enabling dynamic allocation of representation capacity and producing compact yet expressive multi-level representations. Second, we design the RQHC-Transformer, a residual quantized hierarchical causal transformer that decodes motion tokens in a single forward pass. Each transformer block aligns with one quantization level, allowing hierarchical abstraction and temporally coherent generation with strong semantic flow. Compared to diffusion- and LLM-based approaches, MOGO achieves lower inference latency while preserving high motion fidelity. Moreover, its hierarchical latent design enables seamless and controllable infinite-length motion generation, with stable transitions and the ability to adaptively incorporate updated control signals at arbitrary points in time. To further enhance generalization and interpretability, we introduce Textual Condition Alignment (TCA), which leverages large language models with Chain-of-Thought reasoning to bridge the gap between real-world prompts and training data. TCA not only improves zero-shot performance on unseen datasets but also enriches motion comprehension for in-distribution prompts through explicit intent decomposition. Extensive experiments on HumanML3D, KIT-ML, and the unseen CMP dataset demonstrate that MOGO outperforms prior methods in generation quality, inference efficiency, and temporal scalability. Code is available at \url{https://github.com/MiRECoFu/Mogo}.
\end{abstract}

\section{Introduction}
\label{sec:intro}
Text-to-motion generation is a rapidly evolving research area with growing importance in virtual environments, such as gaming, AR/VR, and humanoid robotics~\cite{zhu2023humanmotiongenerationsurvey}. Recent advances have leveraged large language model (LLM)-based techniques to generate high-quality 3D human motion from text descriptions, typically using vector-quantized variational autoencoders (VQ-VAE) and autoregressive decoding strategies~\cite{guo2024momask,zhang2023generating,pinyoanuntapong2024mmmgenerativemaskedmotion,zhong2023attt2m}. Meanwhile, diffusion-based approaches~\cite{Sun2025MCRE, Wang_2025_WACV} have also demonstrated strong generation performance, particularly in terms of motion fidelity and diversity~\cite{tevet2022human}.

Despite these achievements, both diffusion- and LLM-based frameworks face practical limitations~\cite{rombach2022high,chen2023executing,brown2020language,raffel2020exploring}. Diffusion models often rely on iterative refinement processes, which introduce significant inference latency and hinder their suitability for real-time or interactive applications~\cite{rombach2022high}. On the other hand, LLM-based models, although autoregressive, typically involve large parameter sizes and long context dependencies, leading to high memory and computation costs that challenge deployment on lightweight scenarios~\cite{brown2020language}. A broader discussion of related research trends is provided in Section~\ref{sec:relate_work}.

To address the challenges of motion generation quality, inference efficiency, and generalization, we propose MOGO, a unified and efficient transformer-based framework for expressive text-to-motion synthesis. MOGO generates high-fidelity motion sequences from textual input in a single forward pass, combining structural compactness, temporal coherence, and extendable decoding. The framework is composed of four key components, each targeting a core aspect of the generation pipeline:

\begin{itemize}

    \item We propose \textit{MoSA-VQ}, a motion structure-aware quantization framework with two key innovations: 
    (i) a learnable modulation mechanism that adaptively scales residual magnitudes across quantization levels, and 
    (ii) a novel cross-level decorrelation loss that enforces statistical orthogonality between quantized vectors and subsequent residuals. 
    Together, these components enable more compact and hierarchically disentangled representations, leading to better reconstruction under constrained codebook capacity.

    \item \textit{RQHC-Transformer (Residual Quantized Hierarchical Causal Transformer)} functions as the motion decoder. It autoregressively generates multi-layer motion tokens in a single forward pass, with each transformer block dedicated to decoding one level of the quantized hierarchy. This design ensures temporally coherent synthesis while preserving decoding efficiency and compatibility with interactive applications.

    \item \textit{Infinite-Length Generation} is enabled by the autoregressive nature of MOGO’s architecture. Unlike diffusion-based methods that require fixed-length sampling, MOGO supports seamless motion extension with consistent structure and fidelity, making it suitable for real-time or open-ended generation scenarios.

    \item \textit{Textual Condition Alignment (TCA)} leverages large language models with Chain-of-Thought reasoning to bridge the gap between real-world user prompts and the model’s training distribution. Beyond simple style normalization, TCA explicitly decomposes complex motion instructions into structured semantic cues, enabling more accurate motion interpretation. This enhances MOGO's generalization capability across both in-distribution and zero-/few-shot scenarios, while improving control fidelity and robustness to diverse user inputs.

\end{itemize}


\section{Related Work}
\label{sec:relate_work}


\textbf{Human Motion Generation.}  
Recent advances in human motion generation have enabled conditioning on modalities like text, audio, music, and images~\cite{zhu2023humanmotiongenerationsurvey}. Early deterministic models~\cite{ahuja2019language2pose, ghosh2021synthesis} often produced over-smoothed, unrealistic motions. To address this, stochastic approaches, such as GANs~\cite{cai2018deep, wang2020learning} and VAE-based models~\cite{guo2022action2video, petrovich2021action}, improved motion diversity. Text-to-motion generation gained prominence with works like~\cite{guo2022generating, zhou2024emdm, lu2025scamo, fan2025go}, which used temporal VAEs to model text-motion distributions. Recently, diffusion-based methods~\cite{kim2023flame, zhang2024motiondiffuse, zhang2023remodiffuse, kong2023priority, chen2023executing, tevet2022human} and transformer-based approaches~\cite{guo2024momask, zhang2023generating, jiang2023motiongpt} have led the field.

\textbf{LLM-Based Motion Generation Models.}  
LLM-based architectures have become a cornerstone of text-to-motion generation, leveraging their ability to model sequential data and adapt to varied tasks~\cite{guo2024momask, zhang2023generating, pinyoanuntapong2024mmmgenerativemaskedmotion, zhong2023attt2m, jiang2023motiongpt}. Models like MoMask~\cite{guo2024momask} and MMM~\cite{pinyoanuntapong2024mmmgenerativemaskedmotion} use masked token modeling to produce high-quality motions but struggle with extendable output and generalization in low-data or out-of-distribution settings due to their bidirectional design. Conversely, autoregressive models like T2M-GPT~\cite{zhang2023generating} and AttT2M~\cite{zhong2023attt2m} enable sequential generation, making them suitable for real-time applications and scalable with larger datasets, though they often sacrifice some motion quality. Efforts like MotionGPT~\cite{jiang2023motiongpt} integrate multimodal language modeling but face challenges in achieving high-fidelity motion outputs. Our MOGO framework addresses these issues by combining efficient encoding through the MoSA-VQ with a single-pass autoregressive transformer, ensuring both high-quality motion and interactive capabilities.

\textbf{Hierarchical Transformers.}  
Hierarchical transformer architectures excel in domains like NLP~\cite{nawrot2022hierarchical, pappagari2019hierarchical, fang2025and}, image generation~\cite{ding2022cogview2}, and vision tasks~\cite{liu2021Swin, chen2022scaling, 10638169, chen2025hifi}. By processing data at multiple abstraction levels, these models enhance representation capacity and scalability. For instance, Swin Transformers~\cite{liu2021Swin} support scalable high-resolution vision, while CogView2~\cite{ding2022cogview2} enables high-fidelity image synthesis. In motion generation, hierarchical transformers are underexplored, especially for autoregressive frameworks with residual quantization. Our RQHC-Transformer in MOGO leverages hierarchical modeling to efficiently process multi-layer motion tokens, improving quality and generalization.

\textbf{Motion Tokenization.}  
Discretizing continuous motion data into tokens v`ia vector quantization is central to transformer-based motion generation. TM2T~\cite{chuan2022tm2t} introduced VQ-VAE to map motions to discrete sequences. T2M-GPT~\cite{zhang2023generating} enhanced token quality with exponential moving average and codebook reset techniques. AttT2M~\cite{zhong2023attt2m} improved quantization through body-part-aware encoding. MoMask~\cite{guo2024momask} advanced this with residual vector quantization (RVQ), producing multi-level tokens for better reconstruction quality. Compared to prior work, our approach introduces learnable feature scaling into the residual vector quantization process, allowing each quantization level to adaptively adjust the magnitude of its residuals. This scaled RVQ mechanism stabilizes the residual distribution across quantization stages, mitigates the risk of later stages collapsing to noise, and ensures more effective codebook utilization. As a result, the model benefits from improved reconstruction fidelity, better token expressiveness, and enhanced training stability. 

\section{Methods}
\label{sec:methods}

\subsection{Overview of the MOGO Framework}

\begin{figure*}[!t] 
    \centering
    \includegraphics[width=0.88\textwidth]{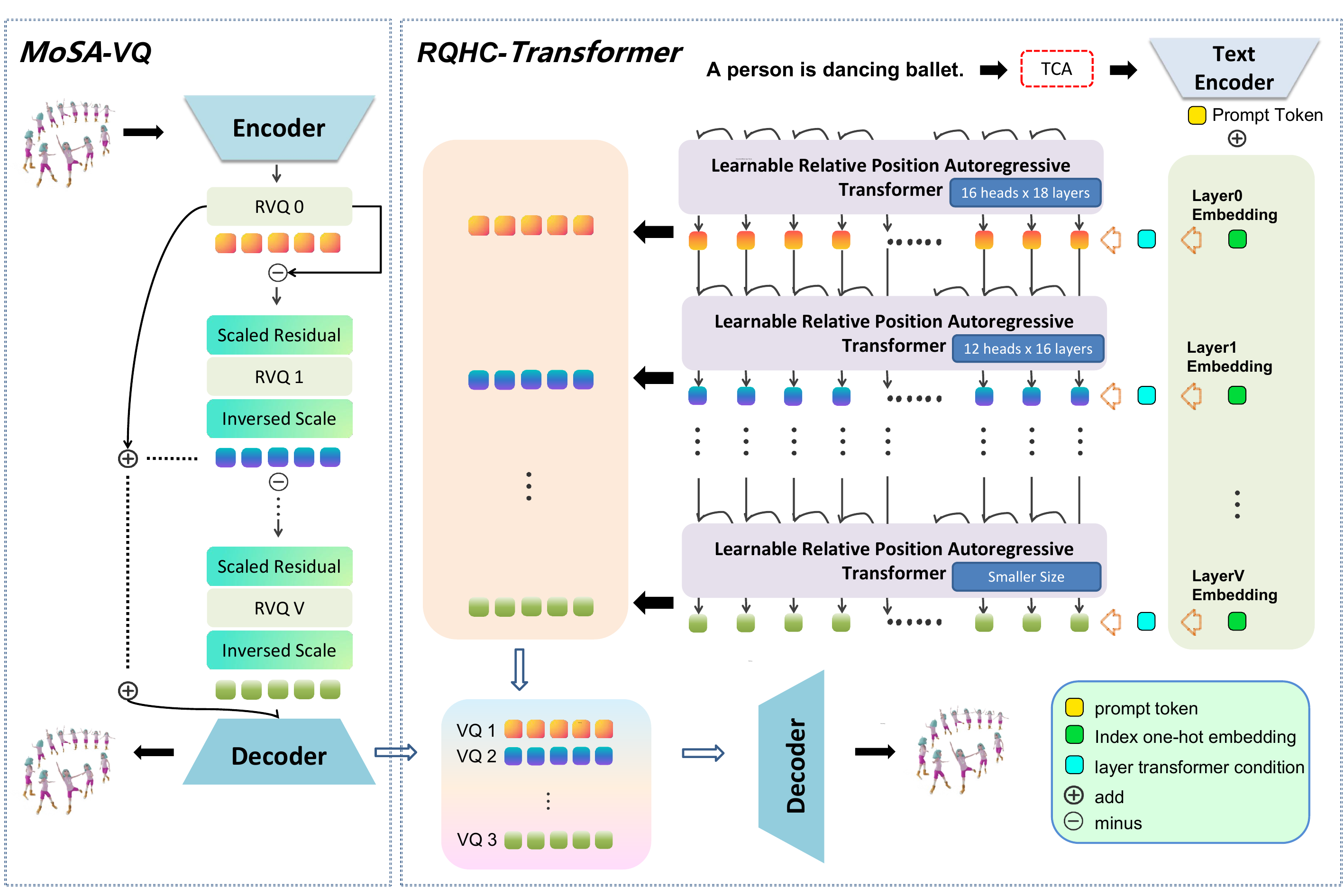}
    \caption{Overview of the proposed MOGO framework. \textit{Left}: We first encode the motion sequence using a hierarchical MoSA-VQ module with learnable feature scaling, indexing a multi-level codebook. \textit{Right}: Then, tokens are autoregressively generated by a RQHC-Transformer with relative positional attention. Finally, the predicted token sequence is quantized and decoded into a full 3D human motion sequence.}

    \label{fig:architechure1} 
\end{figure*}

Our proposed framework, MOGO, as showned in Figure~\ref{fig:architechure1}, consists of four key components designed for efficient and expressive text-to-motion generation. First, MoSA-VQ discretizes motion sequences into multi-level latent tokens using hierarchical residual quantization with adaptive scaling. Second, RQHC-Transformer autoregressively decodes the quantized representations with a stack of level-aligned transformer blocks. Third, MOGO supports infinite-length generation via seamless autoregressive extension. Lastly, we introduce TCA to enhance generalization to real-world prompts. We elaborate on each component in the following subsections.

\subsection{MoSA-VQ: Motion Scale-Adaptive Residual VQ-VAE}
\label{sec:rvqvae}
To transform continuous human motion into discrete representations suitable for autoregressive generation, we build upon residual vector quantized variational autoencoders (RVQ-VAEs) and propose MoSA-VQ. While standard RVQ-based approaches~\cite{guo2024momask, zhong2023attt2m} hierarchically encode motion features through fixed-stage quantization, they often rely on manually tuned feature ranges and static scaling behavior, which limit their adaptability and quantization efficiency—especially in the presence of diverse and complex motions.

\paragraph{Adaptive Hierarchical Quantization.}
To address these limitations, we introduce a learnable feature scaling mechanism into each quantization level. Specifically, each level is equipped with its own learnable scale and bias parameters, which dynamically adjust the amplitude and offset of the input residual features before quantization. This adaptive design allows each level to allocate representational capacity where needed, resulting in more effective codebook usage and higher reconstruction accuracy without the need for handcrafted calibration.

At each hierarchical quantization level \( l \in \{0, 1, \ldots, L\} \), the residual vector is first transformed using a learnable affine mapping. We denote this process as:
\begin{equation}
\mathbf{q}^l = \mathcal{Q}(\mathbf{W}^l \mathbf{r}^l + \mathbf{b}^l), \quad 
\boldsymbol{\phi}^l = (\mathbf{W}^l)^{-1} (\mathbf{q}^l - \mathbf{b}^l),
\end{equation}
where \( \mathbf{W}^l \in \mathbb{R}^{d \times d} \) is a diagonal scaling matrix, \( \mathbf{b}^l \in \mathbb{R}^d \) is a learnable bias vector, \( \mathcal{Q}(\cdot) \) denotes nearest-neighbor quantization via a codebook, and \( \boldsymbol{\phi}^l \) represents the decoded contribution at level \( l \).

This hierarchical quantization scheme enables progressive residual refinement. Unlike conventional residual vector quantization (RVQ) frameworks that apply uniform scaling across all levels, our design introduces level-specific adaptive modulation through learnable parameters \( \mathbf{W}^l \) and \( \mathbf{b}^l \). This flexibility allows coarse levels to capture global structure, while deeper layers focus on subtle motion or spatial details. The inverse affine transformation ensures consistent feature recovery, and stride-1 convolutions in both encoder and decoder maintain temporal continuity without aliasing artifacts. We have
\begin{equation}
\mathbf{r}^{l+1} = \mathbf{r}^l - \boldsymbol{\phi}^l, \quad
\hat{\mathbf{z}} = \sum_{l=0}^{L} \boldsymbol{\phi}^l.
\end{equation}

\paragraph{Training Objective.}
Our full training objective combines four terms: motion reconstruction, codebook commitment, lightweight modulation regularization, and a novel cross-level decorrelation loss.

We adopt an $\ell_1$ reconstruction loss to ensure accurate motion recovery, and a standard commitment loss to encourage the encoder residuals $\mathbf{r}^l$ to remain close to their quantized counterparts $\boldsymbol{\phi}^l$. A mild regularization is also applied to the per-layer scaling and bias parameters $(\mathbf{W}^l, \mathbf{b}^l)$, following prior work on stable quantization with learnable modulation.

Additionally, we introduce a cross-level decorrelation loss, which explicitly penalizes statistical redundancy between the quantized vector $\boldsymbol{\phi}^l$ and the subsequent residual $\mathbf{r}^{l+1}$:
\[
\mathcal{L}_{\text{decor}} = \sum_{l=0}^{L-1} \left\| \mathrm{Cov}\left(\boldsymbol{\phi}^l, \mathbf{r}^{l+1} \right) \right\|_F^2.
\]
This loss is motivated by the principle that, in an ideal residual quantization hierarchy, each level should encode novel information orthogonal to what has already been captured. A nonzero covariance between $\boldsymbol{\phi}^l$ and $\mathbf{r}^{l+1}$ indicates overlapping representation subspaces, leading to inefficient coding and excessive reliance on deeper quantizers. By minimizing their batch-wise covariance, we enforce an approximate statistical orthogonality between levels, which encourages disentangled information flow and hierarchical efficiency.

From an information-theoretic perspective, this decorrelation loss can be viewed as promoting minimal mutual information between consecutive encoding layers, thus maximizing the effective bit utilization of each quantized token. Empirically, we find this regularization yields more compact latent codes and improves reconstruction quality under constrained codebook budgets. The final loss function is:
\begin{align}
\mathcal{L}_{\mathrm{vq}} =\ 
& \gamma \sum_{l=0}^{L} \underbrace{\left( \left\| \mathbf{W}^l \!-\! \mathbf{I} \right\|_F^2 \!+\! \left\| \mathbf{b}^l \right\|_2^2 \right)}_{\text{modulation reg.}} \!+\!
\beta \sum_{l=0}^{L} \underbrace{\left\| \mathbf{r}^l \!-\! \mathrm{sg}[\boldsymbol{\phi}^l] \right\|_2^2}_{\text{commitment}} \nonumber \\
& + \lambda \sum_{l=0}^{L-1} \underbrace{\left\| \mathrm{Cov}\left(\boldsymbol{\phi}^l, \mathbf{r}^{l+1} \right) \right\|_F^2}_{\text{cross-level decorrelation}} +
\underbrace{\| \mathbf{m} - \hat{\mathbf{m}} \|_1}_{\text{reconstruction}}.
\end{align}
Here, $\beta $, $\lambda$ and $\gamma $ are weight factors.

\subsection{RQHC-Transformer: Residual Quantized Hierarchical Causal Transformer}
\label{sec:trainHMT}

To fully leverage the high-quality hierarchical representations produced by MoSA-VQ, we propose the Residual Quantized Hierarchical Causal Transformer (RQHC-Transformer), an autoregressive decoder structurally aligned with the residual quantization process. In contrast to single-resolution or per-layer decoding approaches, RQHC jointly models all quantization levels in a residual-aware and level-aware manner. Moreover, we adopt a causal Transformer design, which—compared to bidirectional transformers, LLM-style decoders, or iterative diffusion models—offers efficient one-pass inference, supports streamable generation, and naturally enables long-horizon synthesis. The coarse-to-fine decoding progression preserves structural semantics, mitigates token interference, and promotes stable, coherent motion generation with improved interpretability.

This hierarchical causal architecture progressively refines coarse motion structures, preserving semantic consistency across abstraction levels and reducing token-level interference. Crucially, its design is structurally aligned with the residual quantization process of MoSA-VQ, enabling stable information flow across layers. Together, this synergy allows MOGO to extend motion from any given frame with temporal continuity and semantic fidelity, laying the foundation for real-time, controllable infinite-length generation.

\paragraph{Hierarchical Causal Generation.} 
To generate the token sequence at quantization level \( l \), we construct an input sequence \( \mathbf{s}^l \) that conditions on both the textual prompt and the residual context from lower levels:
\begin{equation}
\mathbf{s}^l = [\mathbf{p} + \mathbf{q}_{\text{emb}}, \mathbf{t}_{l}^{1:n}],
\end{equation}
where $\mathbf{p}$ is the CLIP-based text prompt embedding, and $\mathbf{q}_{\text{emb}}$ is the embedding representing the current quantization level. The sequence $\mathbf{t}_l^{1:n}$ is computed by aggregating token embeddings across all previous levels at each position:
\begin{equation}
\mathbf{t}_{l}^{1:n} = \left[ \text{Embed}(t_{l}^{1}),\ \text{Embed}(t_{l}^{2}),\ \ldots,\ \text{Embed}(t_{l}^{n}) \right].
\end{equation}
Here, \( n \) denotes the number of motion tokens (i.e., temporal steps) in the sequence. Each \( t_{l}^{i} \) is the \( i \)-th quantized token at layer \( l \), and \( \text{Embed}(\cdot) \) maps each token to its corresponding embedding vector.

\textbf{Relative Positional Encodings.} To effectively handle long motion sequences, we incorporate a relative positional encoding scheme into our causal attention layers. Compared to absolute positional encodings, this approach better preserves attention consistency for varying sequence lengths and enhances the model’s ability to capture long-range dependencies.

Given an input token sequence, the attention score between token $i$ and token $j$ is computed as:
\begin{equation}
\mathbf{A}^{\text{rel}}_{i,j} = \mathbf{q}_i^\top \mathbf{k}_j + \mathbf{q}_i^\top \mathbf{\phi}_{i-j} + u^\top \mathbf{k}_j + v^\top \mathbf{\phi}_{i-j},
\end{equation}
where $\mathbf{q}_i$ and $\mathbf{k}_j$ are the query and key embeddings at position $i$ and $j$, and $\mathbf{\phi}_{i-j}$ is a learned relative position embedding, and $u$ and $v$ are learnable global bias vectors. The final output of the self-attention is:
\begin{equation}
\mathbf{a}_l^n = \text{Softmax}(\mathbf{A}_l^n) \mathbf{V}_l^n,
\end{equation}
where \( \mathbf{A}_l^n \) is the attention score matrix incorporating relative positional encoding at layer \( n \) of the \( l \)-th quantization level. \( \mathbf{V}_l^n \) denotes the value matrix at the same layer.

\paragraph{Training Objective.}
To learn temporally coherent and hierarchically consistent token generation, we adopt a multi-layer autoregressive training objective defined over all quantization levels. The loss function is written as:
\begin{equation}
\mathcal{L}_{\text{ce}} = -\mathbb{E}_{(\mathbf{t}, c) \sim \mathcal{D}} \left[ \sum_{l=1}^{L} \log p_\theta(\mathbf{t}^l \mid \mathbf{t}_{<}^l, c) \right],
\end{equation}
where \( \mathcal{L}_{\text{ce}} \) denotes the total cross-entropy loss computed over all quantization levels. The training data \( \mathcal{D} \) consists of pairs \((\mathbf{t}, c)\), where \( \mathbf{t} = \{ \mathbf{t}^1, \ldots, \mathbf{t}^L \} \) is the set of discrete motion token sequences across \(L\) quantization levels, and \( c \) is the global conditioning signal derived from the input text and the residual context from coarser levels. Each sequence \( \mathbf{t}^l = (t_1^l, t_2^l, \ldots, t_T^l) \) corresponds to the motion tokens at level \( l \), with \( T \) being the number of time steps. The term \( \mathbf{t}_{<}^l \) denotes the causal context of \( \mathbf{t}^l \), i.e., all tokens preceding the current position. The model \( p_\theta \), parameterized by weights \( \theta \), is trained to predict each token autoregressively based on its past tokens and the shared condition \( c \). This objective encourages consistent hierarchical structure and smooth temporal evolution across all levels.

\subsection{Infinite-Length Generation}
\label{sec:infinitegen}

By leveraging MoSA-VQ’s temporally discrete codes and RQHC’s autoregressive decoding, our framework naturally supports infinite-length motion generation from arbitrary time steps. During extension, the model conditions on previously generated tokens (excluding the original prompt embedding), while allowing the text prompt to be substituted dynamically—enabling seamless continuation under updated instructions.

\paragraph{Flexible Continuation with Prompt Switching.}
At any generation step \( n \), given an existing motion token sequence \( \mathbf{t}_l^{1:n-1} = (t_l^1, t_l^2, \ldots, t_l^{n-1}) \) at quantization level \( l \), we continue autoregressive generation from step \( n \) onward using the RQHC-Transformer. To support runtime intervention, we allow prompt switching by replacing the original text condition \( \mathbf{p}_{\text{old}} \) with a new prompt \( \mathbf{p}_{\text{new}} \), yielding an updated conditioning vector \( c = \mathbf{p}_{\text{new}} + \mathbf{q}_{\text{emb}} \).

While the previous tokens \( \mathbf{t}_l^{1:n-1} \) are preserved, the updated prompt influences subsequent decoding. At each step \( n \), RQHC predicts the next token \( t_l^n \) based on the causal context and the new condition:
\begin{equation}
t_l^n \sim p_\theta(t_l^n \mid \mathbf{t}_l^{1:n-1}, c), \quad \forall n > n{-}1.
\end{equation}
This per-step decoding continues autoregressively and can be extended to arbitrary lengths, enabling real-time motion continuation from any point. Crucially, the updated prompt takes effect immediately from step \( n \), without requiring regeneration of prior tokens. Only after completing the desired generation (of any length) do we perform motion decoding to reconstruct full-body trajectories. This design enables flexible, prompt-controllable extension with temporal coherence and efficient computation.

\paragraph{Why This Works.}
This flexible and prompt-adaptive generation capability stems from the close architectural synergy between MoSA-VQ and RQHC-Transformer. MoSA-VQ’s residual quantization provides a structured representation where coarse global motion is captured in early levels and finer dynamics are encoded hierarchically. This layered encoding ensures stability during long-horizon generation, as new content naturally extends existing motion without disrupting overall structure. In advanced, RQHC’s hierarchical causal structure enables prompt substitution at any point by conditioning future generation on preserved prior tokens, ensuring smooth semantic transitions. Together, these properties support real-time extension, runtime editing, and infinite-length motion generation under evolving conditions.


\subsection{Text Condition Alignment for Improved Motion Decoding}
Real-world user prompts for motion generation often deviate from the structured annotations seen during training, varying in style, verbosity, and semantic complexity. Notably, many instructions contain abstract or composite motions that are difficult to execute when treated as a single condition, posing a fundamental challenge to robust generation.

To bridge this gap, we propose TCA, an inference-time strategy that leverages a pretrained large language model \( \mathcal{T} \) to enhance both interpretability and executability of free-form prompts—without modifying model parameters. TCA performs two key functions: (1) \textit{style normalization}, aligning user input with the model’s linguistic prior, and (2) \textit{instruction decomposition}, splitting complex intents into atomic motion steps.

Formally, given an arbitrary user prompt \( \mathbf{c}_{\text{raw}} \in \mathbb{U} \), TCA first produces a normalized instruction \( \mathbf{c}_{\text{norm}} \in \mathbb{S} \), then applies Chain-of-Thought prompting over \( \mathcal{T} \) to decompose it as:
\begin{equation}
[\mathbf{c}^{(1)}, \mathbf{c}^{(2)}, \ldots, \mathbf{c}^{(K)}] = \mathcal{T}(\mathbf{c}_{\text{norm}} \mid \Psi_{\text{CoT}}),
\end{equation}
where \( \Psi_{\text{CoT}} \) encodes few-shot exemplars and reasoning patterns.

Each \( \mathbf{c}^{(k)} \) denotes a semantically atomic sub-instruction, enabling the motion model to synthesize sequences in a stepwise or joint manner with improved controllability, precision, and temporal coherence.

By externally structuring complex instructions into executable units, TCA enhances generation quality across both seen and unseen prompts—without requiring any fine-tuning or additional supervision.

\begin{table*}[t]
  \centering
  \footnotesize
  
  \resizebox{\textwidth}{!}{
  \begin{tabular}{c l c c c c c c}
    \toprule
    \multirow{2}{*}{\textbf{Datasets}} & \multirow{2}{*}{\textbf{Methods}} & \multicolumn{3}{c}{\textbf{R Precision $\uparrow$}} & \multirow{2}{*}{\textbf{FID $\downarrow$}} & \multirow{2}{*}{\textbf{MultiModal Dist $\downarrow$}} & \multirow{2}{*}{\textbf{MultiModality $\uparrow$}} \\
    \cmidrule(lr){3-5}
    & & \textbf{Top 1} & \textbf{Top 2} & \textbf{Top 3} & & & \\
    \midrule

    \multirow{9}{*}{\textbf{HumanML3D}} 
      & MotionDiffuse~\cite{zhang2024motiondiffuse} & 0.491$\pm$0.001 & 0.681$\pm$0.001 & 0.782$\pm$0.001 & 0.630$\pm$0.001 & 3.113$\pm$0.001 & 1.553$\pm$0.042 \\
      & T2M-GPT$^{\dagger}$~\cite{zhang2023generating} & 0.491$\pm$0.003 & 0.680$\pm$0.002 & 0.775$\pm$0.002 & 0.116$\pm$0.004 & 3.118$\pm$0.011 & 1.856$\pm$0.011 \\
      & Fg-T2M~\cite{wang2023fg} & 0.492$\pm$0.002 & 0.683$\pm$0.003 & 0.783$\pm$0.003 & 0.243$\pm$0.019 & 3.109$\pm$0.007 & 1.614$\pm$0.049 \\
      & AttT2M$^{\dagger}$~\cite{zhong2023attt2m} & 0.499$\pm$0.005 & 0.690$\pm$0.006 & 0.786$\pm$0.004 & 0.112$\pm$0.004 & 3.038$\pm$0.016 & \underline{2.452}$\pm$0.043 \\
      & MotionGPT$^{\dagger}$~\cite{jiang2023motiongpt} & 0.492$\pm$0.003 & 0.681$\pm$0.003 & 0.778$\pm$0.002 & 0.232$\pm$0.008 & 3.096$\pm$0.009 & 2.008$\pm$0.084 \\
      & MoMask~\cite{guo2024momask} & \textbf{0.521}$\pm$0.002 & \textbf{0.713}$\pm$0.002 & \textbf{0.807}$\pm$0.002 & \textbf{0.045}$\pm$0.002 & 2.958$\pm$0.008 & 1.241$\pm$0.040 \\
      & MMM~\cite{pinyoanuntapong2024mmmgenerativemaskedmotion} & 0.504$\pm$0.002 & 0.696$\pm$0.003 & 0.794$\pm$0.004 & 0.080$\pm$0.004 & 2.998$\pm$0.007 & 1.226$\pm$0.035 \\
      & MotionAnything~\cite{zhang2025motion} & \fbox{\textbf{0.546}}$\pm$0.002 & \fbox{\textbf{0.735}}$\pm$0.002 & \fbox{\textbf{0.829}}$\pm$0.002 & \fbox{\textbf{0.028}}$\pm$0.001 & \textbf{2.859}$\pm$0.010 & \fbox{\textbf{2.705}}$\pm$0.060 \\
      \cmidrule(lr){2-8}
      & MOGO$^{\dagger}$ & 0.515$\pm$0.003 & 0.709$\pm$0.003 & 0.801$\pm$0.003 & 0.064$\pm$0.002 & 2.951$\pm$0.008 & 2.108$\pm$0.070 \\
      & MOGO with TCA$^{\dagger}$ & \underline{0.527}$\pm$0.007 & \underline{0.722}$\pm$0.008 & \underline{0.827}$\pm$0.012 & \underline{0.038}$\pm$0.003 & \fbox{\textbf{2.849}}$\pm$0.003 & \textbf{2.344}$\pm$0.037 \\

    \midrule

    \multirow{9}{*}{\textbf{KIT-ML}}  
    & MotionDiffuse~\cite{zhang2024motiondiffuse} & 0.417$\pm$0.004 & 0.621$\pm$0.004 & 0.739$\pm$0.004 & 1.954$\pm$0.062 & 2.958$\pm$0.005 & 0.730$\pm$0.013 \\
    & T2M-GPT$^{\dagger}$~\cite{zhang2023generating} & 0.416$\pm$0.006 & 0.627$\pm$0.006 & 0.745$\pm$0.006 & 0.514$\pm$0.029 & 3.007$\pm$0.023 & 1.570$\pm$0.039 \\
    & Fg-T2M~\cite{wang2023fg} & 0.418$\pm$0.005 & 0.626$\pm$0.004 & 0.745$\pm$0.004 & 0.571$\pm$0.047 & 3.114$\pm$0.015 & 1.019$\pm$0.029 \\
    & AttT2M$^{\dagger}$~\cite{zhong2023attt2m} & 0.413$\pm$0.006 & 0.632$\pm$0.006 & 0.751$\pm$0.006 & 0.870$\pm$0.039 & 3.039$\pm$0.016 & \underline{2.281}$\pm$0.043 \\
    & MotionGPT$^{\dagger}$~\cite{jiang2023motiongpt} & 0.366$\pm$0.005 & 0.558$\pm$0.004 & 0.680$\pm$0.005 & 0.510$\pm$0.004 & 3.527$\pm$0.021 & \fbox{\textbf{2.328}}$\pm$0.117 \\
    & MoMask~\cite{guo2024momask} & \textbf{0.433}$\pm$0.007 & \textbf{0.656}$\pm$0.005 & \textbf{0.781}$\pm$0.005 & \textbf{0.204}$\pm$0.011 & \underline{2.779}$\pm$0.022 & 1.131$\pm$0.043 \\
    & MMM~\cite{pinyoanuntapong2024mmmgenerativemaskedmotion} & 0.381$\pm$0.005 & 0.590$\pm$0.006 & 0.718$\pm$0.005 & 0.429$\pm$0.019 & 3.146$\pm$0.019 & 1.105$\pm$0.026 \\
    & MotionAnything~\cite{zhang2025motion} & \fbox{\textbf{0.449}}$\pm$0.007 & \fbox{\textbf{0.678}}$\pm$0.004 & \fbox{\textbf{0.802}}$\pm$0.006 & \fbox{\textbf{0.131}}$\pm$0.003 & \fbox{\textbf{2.705}}$\pm$0.024 & 1.374$\pm$0.060 \\
    \cmidrule(lr){2-8}
    & MOGO$^{\dagger}$ & 0.420$\pm$0.007 & 0.634$\pm$0.007 & 0.754$\pm$0.007 & 0.313$\pm$0.016 & 2.957$\pm$0.029 & 2.063$\pm$0.066 \\
    & MOGO with TCA$^{\dagger}$ & \underline{0.447}$\pm$0.023 & \underline{0.668}$\pm$0.016 & \underline{0.801}$\pm$0.007 & \underline{0.191}$\pm$0.009 & 2.849$\pm$0.007 & \textbf{2.273}$\pm$0.073 \\

    \midrule

    \multirow{7}{*}{\textbf{CMP (zero-shot)}}  
      & T2M-GPT$^{\dagger}$~\cite{zhang2023generating} & 0.061$\pm$0.003 & 0.103$\pm$0.005 & 0.147$\pm$0.006 & 16.092$\pm$0.099 & 4.179$\pm$0.049 & 2.118$\pm$0.033 \\
      & AttT2M$^{\dagger}$~\cite{zhong2023attt2m} & 0.065$\pm$0.004 & 0.109$\pm$0.008 & 0.147$\pm$0.008 & 18.403$\pm$0.071 & \underline{4.048}$\pm$0.017 & 2.208$\pm$0.019 \\
      & MotionGPT$^{\dagger}$~\cite{jiang2023motiongpt} & 0.050$\pm$0.002 & 0.094$\pm$0.002 & 0.133$\pm$0.003 & \underline{15.654}$\pm$0.183 & 4.431$\pm$0.021 & \fbox{\textbf{5.535}}$\pm$0.259 \\
      & MoMask~\cite{guo2024momask} & 0.062$\pm$0.003 & 0.108$\pm$0.005 & 0.150$\pm$0.004 & 24.351$\pm$0.205 & 4.817$\pm$0.022 & 1.651$\pm$0.050 \\
      & MMM~\cite{pinyoanuntapong2024mmmgenerativemaskedmotion} & \underline{0.067}$\pm$0.004 & \underline{0.116}$\pm$0.008 & \underline{0.154}$\pm$0.008 & 17.087$\pm$0.313 & 4.360$\pm$0.017 & 2.802$\pm$0.011 \\
      \cmidrule(lr){2-8}
      & MOGO$^{\dagger}$ & \underline{0.071}$\pm$0.003 & \underline{0.124}$\pm$0.004 & \underline{0.183}$\pm$0.004 & \underline{10.388}$\pm$0.171 & \underline{3.847}$\pm$0.029 & \underline{4.562}$\pm$0.066 \\
      & MOGO with TCA$^{\dagger}$ & \fbox{\textbf{0.122}}$\pm$0.006 & \fbox{\textbf{0.227}}$\pm$0.011 & \fbox{\textbf{0.304}}$\pm$0.004 & \fbox{\textbf{6.873}}$\pm$0.073 & \fbox{\textbf{3.040}}$\pm$0.014 & \textbf{4.299}$\pm$0.013 \\

    \bottomrule
  \end{tabular}
  }
  \caption{Comparison with motion generation models. 
    Bold boxed entries indicate the best performance, bold entries indicate the second-best, and underlined entries indicate the third-best.}
  \label{tab:eval_result_with_GPT}
\end{table*}

\begin{figure*}[h]
    \centering
    \includegraphics[width=0.72\linewidth]{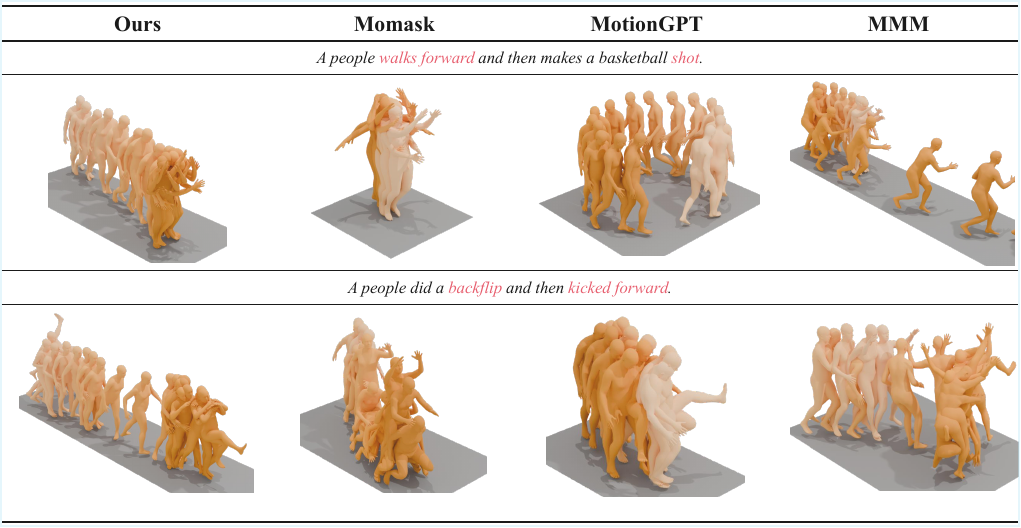}
    \caption{Qualitative comparison of generated motions produced by different models. 
    Compared to prior methods Momask, MotionGPT, and MMM.
    }
    \label{fig:Qualitative experimental on text-to-motion.}
\end{figure*}

\section{Experiments}
\label{sec:experiments}

We comprehensively evaluate our MOGO framework on text-to-motion generation tasks, focusing on motion quality, text-motion alignment, and generalization. Additional experimental results and ablation studies are provided in the appendix for further analysis.

\subsection{Datasets and Evaluation Metrics}
\label{sec:exp_data}

\subsubsection{Datasets} 
We train and evaluate MOGO on two widely used benchmarks, HumanML3D~\cite{guo2022generating} and KIT-ML~\cite{plappert2016kitkit}, following the data splits of T2M~\cite{guo2022generating}. 
To assess out-of-distribution generalization, we further evaluate zero-shot performance on CMP~\cite{yihaoanimationgpt}, which contains non-daily, combat-style motions. 
Prompt engineering is disabled during CMP evaluation to ensure fairness.

\subsubsection{Evaluation Metrics}
We adopt standard metrics from~\cite{guo2022generating}. \textit{FID} (Frechet Inception Distance) measures motion realism and serves as a primary evaluation metric. \textit{R-Precision} evaluates text-motion alignment (Top-1, 2, 3). \textit{MM-Dist} computes the distance between motion and text embeddings. \textit{MultiModality} measures the variance of motions generated from the same text prompt.

\subsubsection{Training Details}
The codebook is sized at $8192 \times 128$ with 6 quantization layers, using stride-1 convolutions and 0.2 dropout. It is trained with AdamW (learning rate $2\times10^{-4}$, batch size 512) for 2000 iterations on an NVIDIA 4090 GPU. RQHC-Transformer   has 6 sub-modules aligned with RVQ layers, with head counts $[16, 12, 6, 2, 2, 2]$ and layer counts $[18, 16, 8, 4, 2, 2]$, and a model dimension of 1024. For HumanML3D, it is trained on an A800-80G with cosine decay from $2.5\times10^{-5}$ to $3\times10^{-6}$, batch size 32, for 1500 epochs. For KIT-ML, a V100-32G is used with a learning rate from $3\times10^{-5}$ to $3\times10^{-6}$, batch size 48.

\subsection{Comparison to State-of-the-art Approaches}
\label{sec:comp_sota}

As shown in Figure~\ref{fig:Qualitative experimental on text-to-motion.}, motions produced by our model are more semantically consistent with the input descriptions and exhibit smoother, more stable transitions without noticeable spatial drift. 
Additional qualitative visualizations, including extended motion generation results, are provided in the Appendix.

\begin{table}[t]
\centering
\scriptsize

\begin{tabular}{llcc}
\toprule
\multicolumn{2}{c}{\textbf{HumanML3D}} & \multicolumn{2}{c}{\textbf{KIT-ML}} \\
\cmidrule(r){1-2} \cmidrule(l){3-4}
\textbf{Method} & \textbf{FID $\downarrow$} & \textbf{Method} & \textbf{FID $\downarrow$} \\
\midrule
M2DM & 0.063$^{\pm 0.001}$ & M2DM & 0.413$^{\pm 0.009}$ \\
T2M-GPT & 0.070$^{\pm 0.001}$ & T2M-GPT & 0.472$^{\pm 0.011}$ \\
MoMask & 0.019$^{\pm 0.001}$ & MoMask & 0.112$^{\pm 0.002}$ \\
MMM & 0.075$^{\pm 0.001}$ & MMM & 0.641$^{\pm 0.014}$ \\
\textbf{MOGO (ours)} & \textbf{0.013}$^{\pm 0.001}$ & \textbf{MOGO (ours)} & \textbf{0.037}$^{\pm 0.001}$ \\
\bottomrule
\end{tabular}
\caption{Comparison of reconstruction quality between our VAE design and prior motion VAEs.}
\label{tab:compare_vae}
\end{table}

\subsubsection{Reconstruction Quality of Motion Encoder}
\label{sec:encoder_recon}

We compare our VAE design with existing motion VAEs—M2DM, T2M-GPT, MoMask, and MMM—on HumanML3D and KIT-ML (Table~\ref{tab:compare_vae}). MOGO achieves the best results on both benchmarks, with an FID of 0.013 on HumanML3D, outperforming MoMask (0.019) and others. On the more challenging KIT-ML dataset, MOGO achieves an FID of 0.037, significantly lower than MoMask (0.112), T2M-GPT (0.472), and MMM (0.641). These results demonstrate the strong reconstruction capability of our VAE across diverse motion distributions.

\subsubsection{Quantitative Results.}
MOGO demonstrates strong performance across HumanML3D, KIT-ML, and zero-shot CMP benchmarks, particularly on FID and R@3. As showned in Table~\ref{tab:eval_result_with_GPT}, 
On HumanML3D, our base model achieves an FID of 0.064, outperforming T2M-GPT (0.116), MotionGPT (0.232), and MotionDiffuse (0.630). 
With TCA, FID slightly improve to 0.038, while R@3 improves from 0.801 to 0.827. 
On KIT-ML, MOGO attains an FID of 0.191 and R@3 of 0.801, competitive with MotionAnything (FID 0.131, R@3 0.802). 
Under the zero-shot CMP setting, FID drops to 10.388 (6.873 with TCA), outperforming MotionGPT (15.654) and T2M-GPT (16.092), while R@3 reaches 0.304, over twice that of the best baseline MMM (0.154). 

Overall, our method consistently achieves top FID and R@3 scores across benchmarks. The TCA module further enhances semantic alignment, especially under distribution shifts, validating the effectiveness of structured condition interpretation.


\subsection{Ablation Study}

\subsubsection{Ablation Study on VAE Depth}
\begin{table}[t]
\centering
\scriptsize

\begin{tabular}{cccc}
\toprule
\textbf{Depth} & \textbf{FID}$\downarrow$& \textbf{R TOP1 $\uparrow$} & MM-Dist$\downarrow$\\
\midrule
1 & 0.070$^{\pm 0.001}$ & 0.502 $^{\pm 0.001}$ & 2.999 $^{\pm 0.006}$\\
3 & 0.021$^{\pm 0.001}$  & 0.508 $^{\pm 0.001}$& 2.992 $^{\pm 0.008}$\\
6 & \textbf{0.016}$^{\pm 0.001}$ & 0.510 $^{\pm 0.001}$ & 2.989 $^{\pm 0.007}$\\
7 & 0.016$^{\pm 0.001}$ & 0.509 $^{\pm 0.001}$ & 2.996 $^{\pm 0.003}$\\
\bottomrule
\end{tabular}

\caption{The impact of different quantization layers on model reconstruction quality when the codebook size is $8192 \times 128$. Bold face indicates the best result.}

\label{tab:compare_vae_kitml}
\end{table}



We evaluate the impact of quantization depth by varying the number of layers while fixing the codebook size to $8192 \times 128$ (Table~\ref{tab:compare_vae_kitml}). A single-layer quantizer performs poorly (FID: 0.070), indicating limited capacity. Increasing to 3 layers significantly improves performance (FID: 0.021), and 6 layers yields the best overall results (FID: \textbf{0.016}, R@1: 0.510, MM-Dist: 2.989). Adding a 7th layer offers no further gain and slightly worsens MM-Dist, suggesting diminishing returns. Overall, 6-layer quantization provides the best balance between expressiveness and generalization. 

\subsubsection{Ablation Study on Layer-Head Configuration}
\begin{table}[t]
\centering
\scriptsize

\begin{tabular}{ccccc}
\toprule
\textbf{Heads} & \textbf{Layers} & \textbf{FID}$\downarrow$ & \textbf{R@1}$\uparrow$ & \textbf{MM-Dist}$\downarrow$ \\
\midrule

\shortstack{[4, 4, 4,\\ 4, 4, 4]} & \shortstack{[6]} & 0.242$^{\pm 0.009}$ & 0.397$^{\pm 0.005}$ & 4.132$^{\pm 0.013}$ \\
\shortstack{[4, 4, 4,\\ 4, 4, 4]} & \shortstack{[6, 6, 6,\\ 6, 6, 6]} & 0.101$^{\pm 0.004}$ & 0.461$^{\pm 0.003}$ & 3.359$^{\pm 0.009}$ \\
\shortstack{[12, 6, 4,\\ 2, 2, 2]} & \shortstack{[16, 8, 6,\\ 4, 2, 2]} & 0.077$^{\pm 0.003}$ & 0.473$^{\pm 0.003}$ & 3.260$^{\pm 0.010}$ \\
\shortstack{[16, 8, 4,\\ 2, 2, 2]} & \shortstack{[18, 10, 6,\\ 4, 2, 2]} & 0.058$^{\pm 0.003}$ & 0.490$^{\pm 0.002}$ & 3.168$^{\pm 0.009}$ \\
\shortstack{[16, 12, 4,\\ 2, 2, 2]} & \shortstack{[18, 16, 6,\\ 4, 2, 2]} & 0.044$^{\pm 0.004}$ & 0.508$^{\pm 0.003}$ & 3.063$^{\pm 0.010}$ \\
\shortstack{[16, 12, 6,\\ 2, 2, 2]} & \shortstack{[18, 16, 8,\\ 4, 2, 2]} & \textbf{0.038}$^{\pm 0.002}$ & \textbf{0.515}$^{\pm 0.003}$ & \textbf{2.951}$^{\pm 0.008}$ \\
\bottomrule
\end{tabular}

\caption{
Impact of model depths and head configurations on HumanML3D results (codebook size: $8192 \times 128$). Bold indicates the best performance.
}
\label{tab:compare_RQHC_kitml}
\end{table}

We study the impact of Transformer depth and attention head configurations on HumanML3D (Table~\ref{tab:compare_RQHC_kitml}). Starting from a uniform setup (\texttt{[4, 4, 4, 4, 4, 4]} layers, \texttt{[6, 6, 6, 6, 6, 6]} heads), the model yields suboptimal results (FID 0.101, MM-Dist 3.359). As we adopt deeper, hierarchically structured layers and progressively narrower heads, performance improves consistently. The best configuration (\texttt{[18, 16, 8, 4, 2, 2]} layers, \texttt{[16, 12, 6, 2, 2, 2]} heads) achieves the lowest FID (\textbf{0.079}), highest R@1 (\textbf{0.505}), and lowest MM-Dist (\textbf{3.002}).

These findings suggest that allocating more depth in early stages and gradually reducing head width enables better modeling of both global semantics and local temporal details. Compared to uniformly scaled variants, hierarchical designs yield stronger R@1 and MM-Dist, highlighting the importance of architectural balance. 

For more ablation details, please check our Appendix.

\section{Conclusion}
\label{sec:conclusion}
We presented MOGO, a one-pass autoregressive framework for high-quality and real-time 3D human motion generation. By combining the MoSA-VQ module for scale-adaptive residual vector quantization with the RQHC-Transformer for hierarchical causal decoding, MOGO achieves efficient and temporally coherent motion synthesis. The integration of TCA mechanism further enhances semantic alignment and enables better generalization in zero- and few-shot scenarios. Our framework supports low-latency, frame-by-frame inference, making it well-suited for real-time and interactive applications. Extensive experiments demonstrate that MOGO consistently outperforms prior transformer-based methods across multiple benchmarks in terms of generation quality and text-motion alignment, establishing a strong foundation for future research on controllable, editable, and interactive motion generation.

\section{Acknowledgements}
This work was supported by Electronics and Telecommunications Research Institute(ETRI) grant funded by the Korean government (24ZC1200, Research on hyper-realistic interaction technology for five senses and emotional experience). The authors would also like to thank Xuan Xu, co-founder of MogoAI, for her valuable discussions and partial computational support.

\bibliography{aaai2026}

\end{document}